%% file: main.tex
\documentclass{article}
\usepackage[preprint]{neurips_2025}
\usepackage[T1]{fontenc}
\usepackage{microtype}
\usepackage{amsmath,amssymb,amsthm}
\usepackage{graphicx}
\usepackage{booktabs}
\usepackage{multirow}
\usepackage{xcolor}
\usepackage{colortbl}
\definecolor{keephi}{HTML}{CFE8D3}  
\definecolor{keeplo}{HTML}{FBF1CF}  
\newcommand{\hi}[1]{\cellcolor{keephi}#1}
\newcommand{\lo}[1]{\cellcolor{keeplo}#1}
\usepackage{enumitem}
\setlist[itemize]{leftmargin=*,itemsep=2pt,topsep=2pt}
\usepackage{caption}
\usepackage{placeins}  
\usepackage[colorlinks,citecolor=brown]{hyperref}
\usepackage[capitalize]{cleveref}

\renewcommand{\paragraph}[1]{\vspace{.1em}\noindent\textbf{#1}}
\graphicspath{{figures/}}

\title{Hybrid Latent Attention for Looped Language Models}
\author{%
  Yuhan Chen$^{1}$, Siyuan Zhang$^{2}$, Nan Wang$^{2}$, Feiyang Kang$^{1}$, Ruoxi Jia$^{1}$\\[4pt]
  $^{1}$Virginia Tech, $^{2}$Independent Researcher%
}

\begin{document}
\maketitle
{\renewcommand{\thefootnote}{}\footnotetext{Correspondence to: \texttt{yuhan@vt.edu}}}

\begin{abstract}
\input{sections/abstract}
\end{abstract}

\input{sections/introduction}
\input{sections/related_work}
\input{sections/method}
\input{sections/training}
\input{sections/experiments}
\FloatBarrier
\input{sections/limitations}
\input{sections/conclusion}

\clearpage
\bibliographystyle{plainnat}
\bibliography{references}

\clearpage
\appendix
\crefalias{section}{appendix}      
\crefalias{subsection}{appendix}
\begin{center}{\Large\bfseries Appendix}\end{center}
\vspace{0.5em}
\input{sections/appendix}

\end{document}

%% file: sections/abstract.tex
Looped language models apply the same stack of layers $T$ times to each token, which deepens the model without adding parameters but multiplies its key--value (KV) cache by $T$.
The larger cache limits how many sequences a GPU can decode at once and slows each decoding step, which reads the whole cache.
We propose Hybrid Latent Attention (HLA), which keeps exact keys and values within a sliding window of $W$ recent tokens and stores each older token as a compact latent that the query of each loop reads directly, without reconstructing keys and values.
We uptrain HLA on Ouro looped models ($T{=}4$) with 1.4B and 2.6B parameters, keeping the pretrained weights frozen and training only the added parameters to reproduce the original attention.
The cache shrinks by $10.7\times$ per token, fitting $4.0$--$8.8\times$ as many concurrent sequences per GPU, and decoding throughput improves by $2.5\times$ at 1K-token contexts and by up to $7.4\times$ at 16K.
HLA retains over 97\% of the original accuracy on math, knowledge and reasoning benchmarks, and 96--100\% on long-context retrieval up to 16K tokens.
After supervised fine-tuning, it performs on par with the fine-tuned original model on competition-level math.

%% file: sections/introduction.tex
\section{Introduction}
\label{sec:intro}

\begin{figure}[t]
\centering
\includegraphics[width=\linewidth]{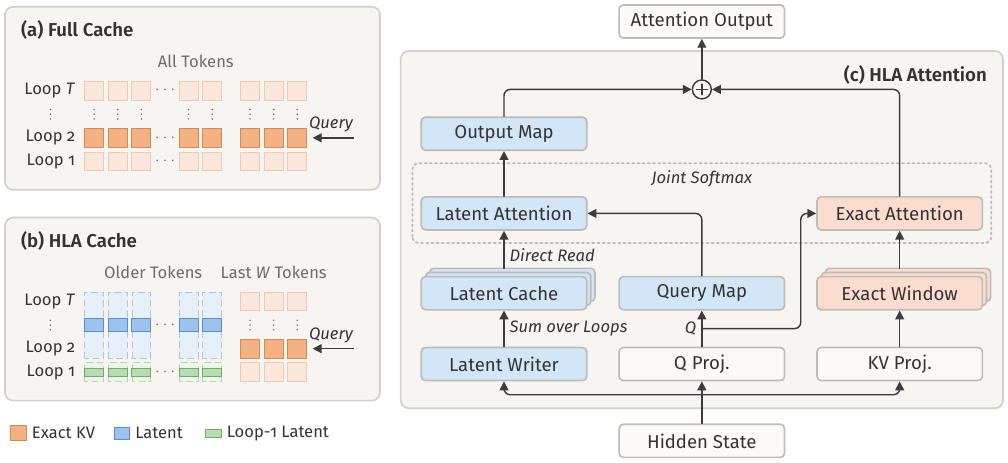}
\caption{Hybrid Latent Attention in one layer of a looped model with $T$ loops. \textbf{(a)} The full cache stores the keys and values (KV) of every loop for every token. \textbf{(b)} HLA keeps exact KV for the last $W$ tokens and stores each older token as one latent shared by loops 2 to $T$ and a smaller latent for loop 1. \textbf{(c)} Attention within one loop. Blue modules are added by HLA and white modules are the frozen pretrained projections. The query and output maps let latent attention read the latent cache directly, and latent and exact attention share one softmax.}
\label{fig:teaser}
\end{figure}

A looped language model applies the same stack of Transformer layers $T$ times to each token, gaining depth without adding parameters~\citep{dehghani2019universal,geiping2025scaling,zhu2025ouro}.
Ouro models with 1.4B and 2.6B parameters and $T{=}4$ are reported to match standard Transformers of up to 12B parameters on a wide range of benchmarks~\citep{zhu2025ouro}.

The parameters saved by looping are paid back at inference time.
Each loop attends to its own keys and values, so the key--value (KV) cache of a looped model is $T$ times that of a non-looped model with the same layers.
This cache bounds how many sequences decode together: an 80\,GB GPU holds only five 16K-token sequences of Ouro-1.4B.
Every decoding step also reads the whole cache, so the time per step of a single sequence grows from 13.6\,ms at 1K tokens of context to 49.3\,ms at 16K (\cref{sec:exp-efficiency}).
Faster decoding therefore requires shrinking both what is stored per token and the work spent on each stored token at every step.

Prior work reduces this cache in two ways.
The first, suggested with Ouro~\citep{zhu2025ouro}, keeps only the last loop's keys and values for generated tokens, so most loops read keys and values that were not computed for them.
Ouro reports a small loss on GSM8K and MATH-500, but \citet{oneill2026lla} find this reuse unsafe, and with long generations a variant at our memory budget retains 60\% of the math accuracy of the full cache on Ouro-1.4B (\cref{sec:experiments}).
The second, Looped Latent Attention (LLA)~\citep{oneill2026lla}, stores a compact latent per token, using the low-rank trajectory of its keys and values across loops, but every decoding step rebuilds the keys and values of each loop from the latents of the whole history; in our measurements LLA reaches 7--18\% of the throughput of the full cache (\cref{sec:exp-efficiency}).

This paper asks whether attention can read such a latent as it is, without reconstruction.
Hybrid Latent Attention (HLA) writes one latent per token and layer as a linear function of the token's hidden states across loops, plus a smaller latent for the first loop, whose keys and values are the hardest to predict (\cref{sec:writer}).
Each loop maps its query into the latent space, scores it against the stored latents, and maps their weighted sum to the attention output (\cref{fig:teaser}); both maps act on the current token only, so no keys or values are rebuilt.

In a model that reads every past token through its latent, the 127 nearest tokens receive about a quarter of the attention but account for about half of its divergence from the original attention (\cref{sec:exp-ablation,sec:app-distance}).
HLA therefore keeps the exact keys and values of the most recent $W$ tokens, a fixed cost per sequence, and reads all earlier tokens through the latent, with one softmax over both parts.

We add HLA to a pretrained model by \emph{uptraining}~\citep{ainslie2023gqa}: with the pretrained weights frozen, the new maps are trained for 1{,}200 steps to match the attention of each layer of the original model.
Fine-tuning is less direct, since a token's latent is complete only after its last loop, yet later tokens read it before their own last loop; a single parallel pass therefore does not reproduce decoding, but two passes, the first writing the latents and the second reading them, come close to it (\cref{sec:finetune}).

On Ouro-1.4B and Ouro-2.6B served with vLLM~\citep{kwon2023vllm}, HLA shrinks the cache by $10.7\times$ per token, from 768 to 72\,KiB on the smaller model.
With the window included, a GPU holds $4.0$--$8.8\times$ as many sequences at contexts from 1K to 16K tokens, and peak decoding throughput rises by $2.5$--$7.4\times$.
With the pretrained weights frozen, HLA retains over 97\% of the original accuracy on MATH500~\citep{hendrycks2021math,lightman2023verify}, MMLU-Pro~\citep{wang2024mmlupro} and BBH~\citep{suzgun2023bbh} and 96--100\% on RULER~\citep{hsieh2024ruler} up to 16K tokens, and after two-pass fine-tuning it performs on par with the fine-tuned original models on AIME and HMMT problems.

Our contributions are as follows.
\begin{itemize}
    \item Hybrid Latent Attention, which keeps exact keys and values for the most recent $W$ tokens and stores older tokens as compact latents that each loop reads directly, without reconstructing keys and values.
    \item Uptraining with frozen weights and two-pass fine-tuning, which add HLA to a pretrained looped model and keep it trainable.
\end{itemize}

%% file: sections/related_work.tex
\section{Related Work}
\label{sec:related}

\paragraph{Looped language models.}
Reusing one block of layers across depth goes back to Universal Transformers~\citep{dehghani2019universal}, ALBERT~\citep{lan2020albert} and deep equilibrium models~\citep{bai2019deq}, and looped Transformers can emulate programs and learn iterative algorithms~\citep{giannou2023looped,yang2024looped}.
Recent work scales looping to language modeling, with a 3.5B-parameter recurrent-depth model~\citep{geiping2025scaling} and the 1.4B and 2.6B Ouro models~\citep{zhu2025ouro}.
These works study quality per parameter and leave open the key--value cache that grows with the number of loops.

\paragraph{Key--value cache reduction.}
Standard Transformers shrink the cache by sharing keys and values across heads~\citep{shazeer2019mqa,ainslie2023gqa} or layers~\citep{sun2024yoco,wu2024lckv,liu2024minicache}, quantizing them~\citep{liu2024kivi,hooper2024kvquant}, or evicting unimportant tokens~\citep{zhang2023h2o,li2024snapkv}, axes largely orthogonal to compression across loops.
Low-rank methods store a latent per token: Palu reconstructs keys and values from it~\citep{chang2025palu}, while MLA absorbs the up-projection into the query and output matrices so that attention reads the latent itself~\citep{deepseek2024v2,meng2025transmla}.
For looped models, Ouro keeps only the last loop's keys and values for generated tokens~\citep{zhu2025ouro}, which LLA~\citep{oneill2026lla} and our long-generation results (\cref{sec:experiments}) find to be an unsafe replacement for the per-loop cache.
LLA, the work closest to ours, stores a per-head low-rank latent of the loop trajectory and, like Palu, reconstructs loop-specific keys and values on every read, which lowers decoding speed from 748 to 108 tokens per second in their report.
HLA instead carries the absorption of MLA across loops with loop-specific maps, so attention reads the latent without reconstruction, and its exact window follows the common practice of keeping recent tokens uncompressed~\citep{rae2020compressive,liu2024kivi}.
FlashLoop~\citep{yang2026flashloop} takes a training-free route through token sparsity, sparse attention and residual quantization, which is complementary in principle to a lower-dimensional latent.

%% file: sections/method.tex
\section{Hybrid Latent Attention}
\label{sec:method}

Hybrid Latent Attention replaces the per-loop keys and values of past tokens with two parts: a latent for every past token, which each loop reads directly, and the exact keys and values of the most recent $W$ tokens.

\subsection{Preliminaries}
\label{sec:prelim}

A looped language model applies its $L$ Transformer layers $T$ times to each token, with the same weights in every loop.
We write $h_{i,t} \in \mathbb{R}^{d}$ for the normalized attention input of token $i$ at loop $t$ in a given layer, and omit the layer index because all layers are treated alike.
Each of the $H$ heads projects $h_{i,t}$ to a query $q_{i,t}$, a key $k_{i,t}$ and a value $v_{i,t}$ in $\mathbb{R}^{d_h}$, with $Hd_h = d$, and rotary embeddings~\citep{su2021roformer} apply a position-dependent rotation $R_i$ to queries and keys.
At loop $t$, token $i$ attends to every token $j \le i$:
\begin{equation}
\begin{gathered}
s_{ij,t} = \frac{(R_i q_{i,t})^\top (R_j k_{j,t})}{\sqrt{d_h}}, \\
a_{i\cdot,t} = \operatorname{softmax}_{j \le i}\,(s_{ij,t}), \qquad
o_{i,t} = \sum_{j \le i} a_{ij,t}\, v_{j,t}.
\end{gathered}
\label{eq:exact}
\end{equation}
The hidden states differ across loops, and so do the keys and values.
Decoding caches the rotated keys $R_j k_{j,t}$ and the values $v_{j,t}$ of every past token at every loop, $2TLd$ numbers per token, which we call the \emph{full cache}.

\subsection{Latent writer}
\label{sec:writer}

The writer replaces the keys and values that a token produces across loops with a latent.
One latent can serve several loops because their keys and values follow a low-rank trajectory across loops~\citep{oneill2026lla}.
The writer therefore takes the hidden states of all the loops a latent covers, and the first loop, whose keys and values are the hardest to predict from the final hidden state, receives a separate, smaller latent (\cref{sec:exp-ablation}).

For each layer, HLA stores two latents per token.
The main latent accumulates a linear map of the hidden state at every loop after the first, and the loop-1 latent is a linear map of the first hidden state:
\begin{equation}
\begin{aligned}
c_j &= \sum_{t=2}^{T} E_t\, h_{j,t} = \begin{bmatrix} c^{K}_j \\ c^{V}_j \end{bmatrix}, \\
c^{(1)}_j &= E_1\, h_{j,1} = \begin{bmatrix} c^{K,1}_j \\ c^{V,1}_j \end{bmatrix}.
\end{aligned}
\label{eq:writer}
\end{equation}
The key part $c^K_j \in \mathbb{R}^{r_k}$ and the value part $c^V_j \in \mathbb{R}^{r_v}$ play the roles of keys and values for loops 2 to $T$, and $c^{K,1}_j, c^{V,1}_j \in \mathbb{R}^{r_1}$ play them for loop 1.
All heads of a layer share these latents.
The main latent is complete once token $j$ has finished its last loop, and it is never updated afterwards.
In total, a token stores $r_k + r_v + 2r_1$ numbers per layer in place of $2Td$.

\subsection{Latent reader}
\label{sec:reader}

Loop $t$ of the current token $i$ reads the latent with two linear maps per head, $A_t \in \mathbb{R}^{d_h \times r_k}$ and $B_t \in \mathbb{R}^{r_v \times d_h}$.
The first carries the query into the key part of the latent space, where it is scored against the stored latents:
\begin{equation}
\hat{s}_{ij,t} = \frac{(\tilde{R}_i A_t^\top q_{i,t})^\top (\tilde{R}_j\, c^{K}_j)}{\sqrt{d_h}}.
\label{eq:latent-score}
\end{equation}
The second carries the attention-weighted sum of value latents back to the head dimension, as given in \cref{eq:hybrid} below.
Loop 1 uses its own maps $A_1, B_1$ on $c^{K,1}_j$ and $c^{V,1}_j$ in the same way.
$A_t$ acts on one query and $B_t$ on one summed vector, so their cost does not depend on the number of past tokens, and no per-loop key or value is formed for any of them.

\Cref{eq:latent-score} needs a rotation $\tilde{R}_j$ that acts on latents.
In \cref{eq:exact} the rotation $R_j$ turns each pair of key coordinates by an angle $j\theta_f$, with one frequency $\theta_f$ per pair.
We organize the key latent in pairs as well, give every frequency the same number $m = r_k/d_h$ of them, and let $\tilde{R}_j$ turn a latent pair of frequency $f$ by $j\theta_f$.
The stored key latent is rotated once, when it is written.
This rotation commutes with any linear combination of key pairs that share a frequency, which TransMLA uses to combine keys across heads~\citep{meng2025transmla}; here the combination also runs across loops.
A latent of this form keeps $m$ of the $(T-1)H$ key directions of each frequency, so its scores approximate the original ones and are exact only when $m = (T-1)H$.
Our initialization takes the leading principal directions of each frequency (\cref{sec:uptraining}), and \cref{sec:app-init} gives the construction in full.
After training, $A_t$ is a dense matrix and this property is no longer guaranteed (\cref{sec:app-rope}); we instead train with shifted positions so that the learned scores remain accurate at long distances (\cref{sec:app-shift}).

\subsection{Exact sliding window}
\label{sec:window}

We find that the most recent tokens account for a large share of the error of the latent (\cref{sec:exp-ablation}), so we store them exactly.
HLA keeps the exact keys and values of the $W$ most recent tokens and reads the latent only for tokens further back.
Token $i$ at loop $t$ forms one score vector from the latent scores $\hat{s}_{ij,t}$ of tokens $j < i-W$ and the exact scores $s_{ij,t}$ of tokens $i-W \le j \le i$, normalizes it with a single softmax, and combines the two kinds of values:
\begin{equation}
o_{i,t} = B_t^\top \sum_{j < i-W} a_{ij,t}\, c^{V}_j \;+ \sum_{i-W \le j \le i} a_{ij,t}\, v_{j,t}.
\label{eq:hybrid}
\end{equation}
The token's own keys and values are always exact, since they are computed in the current step.
With $W{=}0$, \cref{eq:hybrid} reduces to pure latent attention.

A sequence of $n$ tokens thus stores the latent for every token and the full cache for $W$ of them.
The latent takes $L(r_k + r_v + 2r_1)$ numbers per token, and the window adds $2TLdW$ numbers per sequence, independent of $n$.
As $n$ grows, the total approaches a fraction $(r_k + r_v + 2r_1)/(2Td)$ of the full cache.

During inference, a prompt is processed in one pass with exact attention, as in the original model, and the latents of its tokens are written from that pass.
Generation then proceeds one token at a time with \cref{eq:hybrid}.
Each generated token writes its latent as it finishes its loops and places its exact keys and values in the window, from which the oldest token is dropped.
A token is thus read exactly while it is among the $W$ most recent and through its latent afterwards.

%% file: sections/training.tex
\section{Training}
\label{sec:training}

\subsection{Uptraining}
\label{sec:uptraining}

HLA adds the maps $E_t$, $A_t$ and $B_t$ to a pretrained model, 0.35B parameters for Ouro-1.4B and 0.70B for Ouro-2.6B, and uptraining trains only these maps with the pretrained weights frozen.

We initialize $E_t$, $A_t$ and $B_t$ with a principal component analysis of the keys and values that the pretrained model produces on 128 calibration sequences: the writer projects them onto the leading components, and the reader maps queries and outputs through the same components (\cref{sec:app-init}).
For the keys, the components are computed separately for each rotary frequency, as in TransMLA~\citep{meng2025transmla}, but across loops 2 to $T$ as well as across heads, which gives the frequency-aligned form required in \cref{sec:reader}.
The values of loops 2 to $T$ are projected jointly, and loop 1 is treated on its own.
At full rank this initialization reproduces the original attention exactly; at the ranks we use it only approximates it, and uptraining reduces the remaining error.

The pretrained model with its full cache serves as the reference.
We run it on a training sequence and record, for every layer and loop, its attention distribution $a^\star_{i\cdot,t}$ and its attention output $o^\star_{i,t}$ after the output projection.
From the same hidden states we compute the HLA distribution $\hat{a}_{i\cdot,t}$ and output $\hat{o}_{i,t}$ with \cref{eq:latent-score,eq:hybrid}, and minimize
\begin{equation}
\begin{aligned}
\mathcal{L} = \frac{1}{LT} \sum_{\text{layers}} \sum_{t=1}^{T}
\Bigl[ &\operatorname{KL}\!\left(a^\star_{i\cdot,t} \,\middle\|\, \hat{a}_{i\cdot,t}\right) \\
&+ \frac{\lVert \hat{o}_{i,t} - o^\star_{i,t} \rVert^2}{\lVert o^\star_{i,t} \rVert^2} \Bigr],
\end{aligned}
\label{eq:loss}
\end{equation}
averaged over tokens and heads.
The first term fits where each loop attends, and the second fits what it reads.
Scores inside the window are exact in both models, so the latent is supervised only on the tokens it serves at decoding time.

One forward pass of the reference provides the hidden states of every token, layer and loop, so the loss is evaluated for all of them in parallel and each layer is fitted independently.

Uptraining has two phases and uses the same settings for both models.
The first runs 1{,}000 steps on sequences of up to 2{,}048 tokens, drawn from the math reasoning traces of OpenR1-Math-220k~\citep{openr1math2025} and the web text of FineWeb-Edu~\citep{penedo2024fineweb}.
Such sequences never show the latent a distant token, so for three quarters of them we shift the positions of up to four chunks by random offsets, as in PoSE~\citep{zhu2024pose}, which exposes the maps to relative distances of up to 16K tokens without longer inputs.
The second phase runs 200 steps and adds eight documents of 16K tokens to every batch, taken from the books and code repositories of the ProLong training data~\citep{gao2025prolong}, so that the latent is also fitted on real long contexts.
Data and optimizer settings are detailed in \cref{sec:appendix}.

\subsection{Fine-tuning}
\label{sec:finetune}

To test whether the latent supports further training, we fine-tune all weights, including the HLA maps.
Uptraining runs HLA attention on the hidden states of the reference, so it never sees the error that HLA attention accumulates during generation; the accuracy in \cref{sec:exp-quality} is measured by decoding and includes it.
Fine-tuning minimizes the task loss of the HLA model itself, so its forward pass has to run HLA attention on the model's own hidden states, as decoding does, which raises a difficulty that the full cache does not have.

With the full cache, loop $t$ of token $i$ reads the loop-$t$ keys and values of earlier tokens, so all tokens of a training sequence can advance through loop $t$ together.
With HLA, loop 2 of token $i$ already reads the main latent of every token $j < i-W$, and \cref{eq:writer} completes that latent only after loop $T$ of token $j$.
Decoding meets this order by generating one token at a time, but processing a training sequence token by token would give up the parallelism of training.

We recover parallel training with two passes over the sequence.
The first pass runs the model with exact attention and writes the main latents of all tokens with \cref{eq:writer}.
The second pass runs all tokens in parallel under the decoding rule of \cref{eq:hybrid}: prompt tokens attend exactly, as at inference, and response tokens read exact keys and values inside the window and, beyond it, loop-1 latents from the same pass and main latents from the first pass.
The training loss is computed on the second pass, and gradients flow through both passes.
For prompt tokens, the first-pass latents equal those of decoding; for response tokens, they come from hidden states computed with exact attention rather than with HLA.
Rewriting the main latents from the second pass and running it again makes $W{+}1$ more response tokens match decoding exactly, and two passes already come close to sequential decoding (\cref{sec:app-twopass}).

%% file: sections/experiments.tex
\section{Experiments}
\label{sec:experiments}

We evaluate HLA on three questions: how much memory and decoding time it saves (\cref{sec:exp-efficiency}), how much of the original attention and accuracy it keeps after uptraining (\cref{sec:exp-curves}), and whether it can still be fine-tuned (\cref{sec:exp-finetune}).
\Cref{sec:exp-ablation} then examines the design choices.

\begin{figure}[t]
\centering
\includegraphics[width=\linewidth]{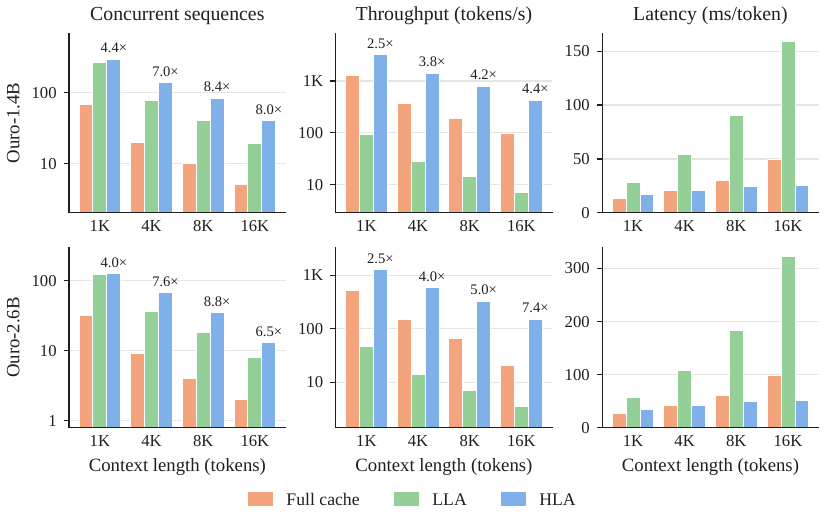}
\caption{Decoding on one A100 GPU with the full cache, LLA and HLA; each sequence has a prompt of the given length and generates 256 tokens. Left: the largest batch that fits in memory. Middle: generated tokens per second summed over the batch, at the best batch size. Right: time per generated token for a single sequence. Labels give the ratio of HLA to the full cache; the left and middle columns use logarithmic axes. \Cref{tab:efficiency} lists the values.}
\label{fig:efficiency}
\end{figure}

\subsection{Setup}
\label{sec:exp-setup}

\paragraph{Models.}
We use the pretrained Ouro-1.4B and Ouro-2.6B checkpoints~\citep{zhu2025ouro} with $T{=}4$ loops.
Both have hidden size $d{=}2048$; Ouro-1.4B has $L{=}24$ layers and Ouro-2.6B has $L{=}48$.
HLA uses $r_k = r_v = 512$, $r_1 = 256$ and a window of $W{=}128$ for both models, and is uptrained as described in \cref{sec:uptraining}.

\paragraph{Cache size.}
At 16-bit precision the full cache holds $2TLd$ numbers per token, 768\,KiB on Ouro-1.4B and 1{,}536\,KiB on Ouro-2.6B.
HLA stores $r_k + r_v + 2r_1 = 1{,}536$ numbers per layer instead of $2Td = 16{,}384$, which is 72 and 144\,KiB per token, $10.7\times$ less.
The window adds 96 and 192\,MiB per sequence, so a 16K-token sequence takes $9.8\times$ less memory than with the full cache.

\paragraph{Baselines.}
The \emph{full cache} is the original model.
\emph{Single-loop sharing} keeps the full cache of the last loop only: every earlier loop reads its own keys and values inside the same window of $W{=}128$ tokens and those of the last loop beyond it.
It needs no training and stores a quarter of the full cache per token, which is $2.7\times$ the latent of HLA.
\emph{LLA}~\citep{oneill2026lla} stores a latent of rank 256 per head, as much as single-loop sharing, and rebuilds the keys and values of every loop from it at each step; for accuracy we distill its codec with its authors' objective for 1{,}000 steps and decode with its two-pass decoder (\cref{sec:app-eval}).

\paragraph{Benchmarks.}
We measure math problem solving on MATH500~\citep{hendrycks2021math,lightman2023verify}, knowledge on MMLU-Pro~\citep{wang2024mmlupro}, reasoning on BBH~\citep{suzgun2023bbh}, and long-context retrieval on RULER~\citep{hsieh2024ruler} at context lengths of 4K, 8K and 16K tokens.
MMLU-Pro, BBH and RULER are run with lm-evaluation-harness~\citep{biderman2024lmeval}, except for the two question-answering tasks of RULER.
With their default prompt and 32 new tokens, Ouro often begins by reasoning about the question and runs out of tokens before it answers, so the score reflects how often a cache leads into such a reply more than how well it retrieves (\cref{sec:app-eval}).
We therefore wrap these prompts in the chat template of the model and allow 2{,}048 new tokens; the other 11 tasks begin the answer in the prompt and keep their settings.
For MATH500 we sample four solutions per problem, with up to 8{,}192 new tokens each, and report mean accuracy.
Fine-tuned models are evaluated on 90 competition problems from AIME 2024~\citep{aime2024hf}, AIME 2025~\citep{aime2025hf} and HMMT February 2025~\citep{balunovic2025matharena}, with 16 samples per problem.

\paragraph{Serving.}
All models are served with vLLM~\citep{kwon2023vllm} at 16-bit precision on A100 GPUs with 80\,GB of memory.
HLA is implemented as an attention backend that stores the latent and the window in vLLM's paged cache (\cref{sec:app-serving}).

\begin{figure}[t]
\centering
\includegraphics[width=\linewidth]{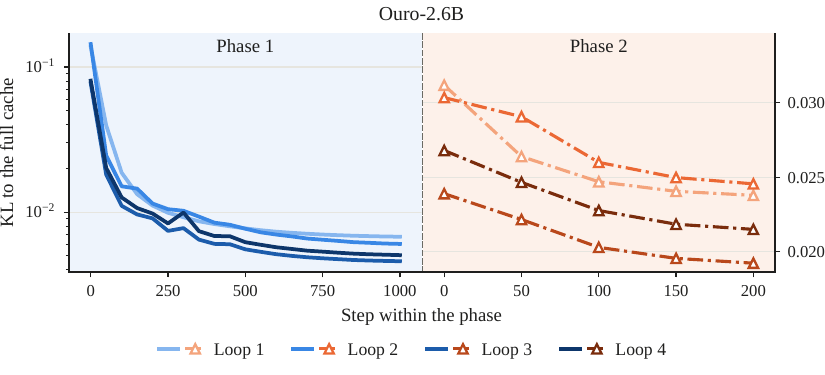}
\caption{Uptraining curves on held-out data for Ouro-2.6B (Ouro-1.4B in \cref{fig:curves-14}): KL divergence between the attention of HLA and of the full cache at each loop. Phase 1 is evaluated on short sequences (logarithmic axis), phase 2 on 16K-token documents (linear axis, right).}
\label{fig:curves}
\end{figure}

\subsection{Memory and decoding speed}
\label{sec:exp-efficiency}

At 1K tokens of context HLA fits about four times as many sequences on one GPU as the full cache, and at 4K to 16K between 6.5 and 8.8 times as many (\cref{fig:efficiency}).
The ratio is lower at 1K because the window costs every sequence the same amount and weighs more when the sequence is short.
Throughput rises less than capacity: by 2.5 times at 1K for both models, and by 4.4 and 7.4 times at 16K.

The time of one decoding step explains part of this gap.
For a single Ouro-1.4B sequence, the latency of the full cache grows by about 2.4\,ms per thousand tokens of context, from 13.6\,ms at 1K to 49.3\,ms at 16K, and that of HLA by about 0.5\,ms, from 17.1 to 25.1\,ms.
HLA is thus 26\% slower at 1K, equal at 4K and twice as fast at 16K, and Ouro-2.6B behaves alike.
We attribute its higher fixed cost to the maps $A_t$ and $B_t$ and to attending to the window and the latent separately.

LLA shows what reconstruction costs.
We serve its reconstruct path in the same engine; the timed codec is not distilled, since its values do not affect decoding time.
LLA fits about four times as many sequences as the full cache, yet the larger batch barely helps: on Ouro-1.4B at 16K it produces 6.3 tokens per second for one sequence and 7.2 for 19.
Rebuilding keys and values for the whole history thus appears to dominate each step, and its peak throughput stays at 7 to 18\% of that of the full cache, while HLA, which reads its latent directly, reaches 28 to 59 times the throughput of LLA.

Fixed-length prompts leave out contexts that grow during generation.
We therefore let the fine-tuned models of \cref{sec:exp-finetune} generate 16 solutions to each of 30 competition problems until completion, on four GPUs per method with the largest batch that each cache allows.
HLA produces 1{,}176 tokens per second per GPU on Ouro-1.4B and 492 on Ouro-2.6B, 3.8 and 4.3 times the 312 and 114 of the full cache.

\subsection{Uptraining}
\label{sec:exp-curves}
\label{sec:exp-quality}

The savings above come from replacing the cache that the model was pretrained with, so we next measure how much of its behavior HLA keeps after uptraining.
The divergence that uptraining minimizes is measured on held-out data and computed on the hidden states of the reference, as in training (\cref{fig:curves,fig:curves-14}).
At initialization the attention of HLA differs from the original by 0.13 nats on average for Ouro-1.4B and 0.11 for Ouro-2.6B.
The first phase reduces this to 0.006 for both models, most of it in the first 200 steps, and loops 1 and 2 remain the hardest to fit, in line with \cref{tab:observations}.
The second phase lowers the divergence on 16K-token documents by about a fifth for both models and raises it slightly on short sequences, from 0.0063 to 0.0067 for Ouro-1.4B.

\begin{table}[t]
\centering
\caption{Accuracy (\%) with the pretrained weights frozen. Single-loop sharing and HLA also keep exact keys and values for the last 128 tokens (96\,MiB per sequence on Ouro-1.4B, 192\,MiB on Ouro-2.6B), not counted in the per-token cache. Shading marks the accuracy that HLA retains relative to the full cache: at least 97\% (green) and 90--97\% (yellow). Bold marks HLA results above LLA. $^\dagger$One sample per problem instead of four, because LLA decodes slowly.}
\label{tab:quality}
\vspace{2pt}
\small
\setlength{\tabcolsep}{5pt}
\begin{tabular}{llrcccccc}
\toprule
 & & Cache per & \multicolumn{3}{c}{RULER} & & & \\
\cmidrule(lr){4-6}
Model & Method & token (KiB) & 4K & 8K & 16K & MMLU-Pro & BBH & MATH500 \\
\midrule
\multirow{4}{*}{Ouro-1.4B}
 & Full cache          &   768 & 89.5 & 85.2 & 77.5 & 48.9 & 71.0 & 75.9 \\
 & Single-loop sharing &   192 & 19.0 & 10.0 & 7.0 & 16.1 & 43.3 & 45.9 \\
 & LLA                 &   192 & 87.4 & 82.7 & 71.1 & 46.6 & 69.1 & 73.4$^\dagger$ \\
 & HLA                 &    72 & \hi{\textbf{88.7}} & \hi{\textbf{84.2}} & \lo{\textbf{74.4}} & \hi{\textbf{47.7}} & \hi{\textbf{69.4}} & \hi{\textbf{74.1}} \\
\midrule
\multirow{4}{*}{Ouro-2.6B}
 & Full cache          & 1{,}536 & 92.3 & 86.3 & 82.9 & 56.0 & 80.1 & 81.1 \\
 & Single-loop sharing &   384 & 71.4 & 64.5 & 52.6 & 52.6 & 75.8 & 73.9 \\
 & LLA                 &   384 & 92.5 & 85.4 & 81.0 & 54.2 & 79.3 & 80.6$^\dagger$ \\
 & HLA                 &   144 & \hi{92.2} & \hi{85.0} & \hi{\textbf{81.5}} & \hi{\textbf{55.5}} & \hi{\textbf{79.8}} & \hi{79.4} \\
\bottomrule
\end{tabular}
\end{table}

With the pretrained weights unchanged, HLA stays close to the full cache (\cref{tab:quality}).
On MATH500, MMLU-Pro and BBH it loses 0.3 to 1.8 points and retains 97.5 to 99.6\% of the original accuracy.
On RULER the loss grows with the context length, from 0.8 points at 4K to 3.1 at 16K for Ouro-1.4B, where it retains 96.0\%, and from 0.1 to 1.4 points for Ouro-2.6B.
Retrieval separates configurations more than math does (\cref{sec:app-sizes}), so we chose the window, shifted positions and long-document phase on RULER.

Single-loop sharing stores more than HLA and loses far more.
On Ouro-1.4B its MATH500 accuracy drops from 75.9 to 45.9, with 44\% of its generations reaching the length limit against 12.5\% for HLA, and its RULER accuracy is at most 19\%.
Ouro-2.6B tolerates sharing better but still loses 7.2 points on MATH500 and 30.3 on RULER at 16K.
The keys and values of one loop thus do not stand in for those of the others, whereas a latent written from all loops does at a smaller size.

LLA stores 2.7 times as much per token as HLA and, on Ouro-1.4B, is less accurate on every benchmark, by 3.3 points on RULER at 16K and 1.1 on MMLU-Pro.
On Ouro-2.6B the two stay within 1.3 points of each other, LLA ahead on MATH500 and on RULER at 4K and 8K, HLA ahead at 16K and on MMLU-Pro and BBH.
Rebuilding the keys and values thus brings no consistent gain in accuracy over reading the latent directly, while it lowers throughput by a factor of 28 to 59.

\subsection{Fine-tuning}
\label{sec:exp-finetune}

We fine-tune each model on math reasoning traces for 400 steps with a batch of 128 sequences and a learning rate of $10^{-5}$, which is about one pass over the data.
The original model is fine-tuned in the standard way.
The HLA model starts from the uptrained checkpoint and is fine-tuned with the two-pass procedure of \cref{sec:finetune}, updating all weights.
Ouro-1.4B uses one repetition of the second pass and Ouro-2.6B none (\cref{sec:app-twopass}).
Measured every 50 steps, the validation losses of the two stay within 0.006 of each other and end at 0.293 and 0.292 for Ouro-1.4B and at 0.267 and 0.269 for Ouro-2.6B.

\begin{table}[!ht]
\centering
\caption{Accuracy (\%) on competition math after fine-tuning on the same data for the same number of steps, averaged over 16 samples per problem. The better value of each pair is in bold.}
\label{tab:finetune}
\vspace{2pt}
\small
\setlength{\tabcolsep}{4pt}
\begin{tabular}{llcccc}
\toprule
& & \multicolumn{2}{c}{AIME} & HMMT & \\
\cmidrule(lr){3-4} \cmidrule(lr){5-5}
Model & Method & 2024 & 2025 & 2025 & Avg. \\
\midrule
\multirow{2}{*}{Ouro-1.4B}
 & Full cache & \textbf{20.6} & 19.8 &  8.1 & 16.2 \\
 & HLA        & 19.2 & \textbf{20.2} & \textbf{9.8} & \textbf{16.4} \\
\midrule
\multirow{2}{*}{Ouro-2.6B}
 & Full cache & 35.4 & 30.4 & 15.6 & 27.2 \\
 & HLA        & \textbf{35.6} & \textbf{31.9} & \textbf{16.5} & \textbf{28.0} \\
\bottomrule
\end{tabular}
\end{table}

\Cref{tab:finetune} gives the accuracy of the fine-tuned models.
HLA reaches 16.4\% against 16.2\% for the full cache on Ouro-1.4B, and 28.0\% against 27.2\% on Ouro-2.6B, so its average is above that of the fine-tuned original at both sizes.
It is also ahead on five of the six pairs of model and competition.
These margins are within noise: paired bootstrap intervals over the 90 problems are $[-1.6, +2.1]$ and $[-1.2, +2.9]$ points for the two averages, and no difference on a single competition exceeds 1.7 points.
The latent cache is thus fully competitive with the full cache after fine-tuning, at a tenth of its size per token.

\subsection{Ablations}
\label{sec:exp-ablation}

We first test the inputs of the writer on the keys and values of the original Ouro-1.4B, and then ablate the exact window, shifted positions and the long-document phase, each by comparing two Ouro-1.4B models with frozen weights that differ in that component only.

The writer of \cref{sec:writer} reads the hidden states of every loop that a latent covers and gives loop 1 a separate latent; \cref{tab:observations} gives the measurements behind both choices.
A ridge regression from the last hidden state $h_{j,4}$ predicts the keys of loop 3 with $R^2 = 0.93$, those of loop 2 with 0.80 and those of loop 1 with 0.66, and the values less well, so $h_{j,4}$ alone loses part of the keys and values of earlier loops.
A rank-512 projection fitted to loops 2--4 jointly keeps 0.79 of the key variance and 0.89 of the value variance, against 0.80 and 0.93 for a single loop, so one latent covers three loops at about the rank of one.
Loop 1, the least predictable, receives its own smaller latent so that it does not enlarge the shared one.

\begin{table}[!ht]
\centering
\caption{Keys and values of the original Ouro-1.4B, averaged over the 24 layers. Left: $R^2$ of a ridge regression from the hidden state of the last loop to the keys and values of each loop. Right: fraction of variance kept by a rank-512 projection fitted to one loop or to loops 2--4 jointly.}
\label{tab:observations}
\vspace{2pt}
\small
\setlength{\tabcolsep}{4pt}
\begin{tabular}{lcccccc}
\toprule
 & \multicolumn{4}{c}{$R^2$ from $h_{j,4}$} & \multicolumn{2}{c}{Var.\ at rank 512} \\
\cmidrule(lr){2-5} \cmidrule(lr){6-7}
Loop & 1 & 2 & 3 & 4 & One & 2--4 \\
\midrule
Keys   & 0.66 & 0.80 & 0.93 & 1.00 & 0.80 & 0.79 \\
Values & 0.55 & 0.71 & 0.89 & 1.00 & 0.93 & 0.89 \\
\bottomrule
\end{tabular}
\end{table}

\begin{figure}[!ht]
\centering
\includegraphics[width=\linewidth]{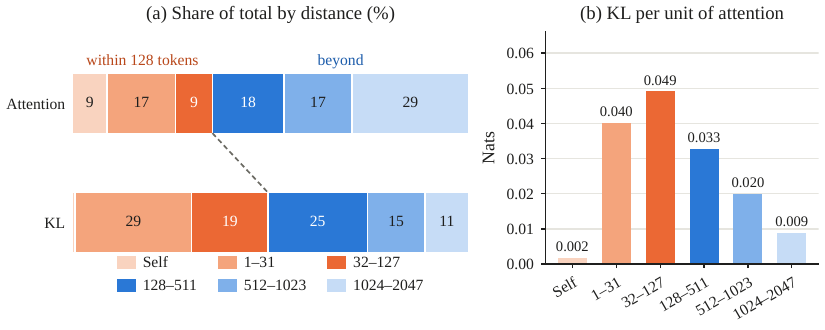}
\caption{Attention divergence of HLA without a window ($W{=}0$) on Ouro-1.4B, by distance between the query and the attended token. \textbf{(a)} How the original attention (top) and the KL divergence to it (bottom) are distributed over six ranges of distance; each bar sums to 100\%. Orange ranges lie within the window of $W{=}128$ tokens used in the main experiments, blue ranges beyond it. \textbf{(b)} Divergence per unit of original attention in each range.}
\label{fig:distance}
\end{figure}

The variant without a window sets $W{=}0$ in uptraining and serving, goes through the same two phases and stores the same latent of 72\,KiB per token.
It loses 8.1 points on MATH500, 7.9 on MMLU-Pro, 8.4 on BBH and 11.5 on RULER at 16K (\cref{tab:ablation}).
The window costs 96\,MiB per sequence, as much as the latent of 1{,}365 tokens, and recovers most of the gap to the full cache.
Splitting the divergence of the model without a window by the distance to the attended token shows where this gap arises (\cref{fig:distance}, \cref{sec:app-distance}).
Tokens at distances 1 to 127 hold 26.6\% of the original attention but 48.7\% of the divergence.
Per unit of attention, the divergence is 0.040 and 0.049 nats at distances 1--31 and 32--127 and 0.009 beyond 1{,}024, so a window of 128 tokens replaces the latent where it is least accurate.

\begin{table}[!ht]
\centering
\caption{Effect of the exact window on Ouro-1.4B with frozen weights. Both variants are uptrained with the full recipe and store a latent of 72\,KiB per token. RULER uses the original settings, unlike \cref{tab:quality} (\cref{sec:app-eval}).}
\label{tab:ablation}
\vspace{2pt}
\small
\setlength{\tabcolsep}{2.5pt}
\begin{tabular}{lcccccc}
\toprule
 & \multicolumn{3}{c}{RULER} & & & \\
\cmidrule(lr){2-4}
$W$ & 4K & 8K & 16K & MMLU-Pro & BBH & MATH500 \\
\midrule
0   & 78.3 & 71.9 & 57.6 & 39.8 & 61.0 & 66.0 \\
128 & 83.2 & 78.7 & 69.1 & 47.7 & 69.4 & 74.1 \\
\bottomrule
\end{tabular}
\end{table}

\begin{figure}[!ht]
\centering
\includegraphics[width=\linewidth]{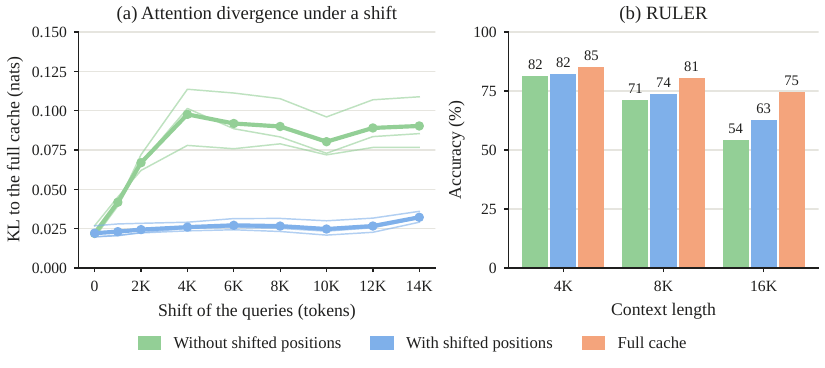}
\caption{Shifted positions on an earlier HLA model with $W{=}32$ on Ouro-1.4B, continued for 100 steps without (green) and with (blue) them. \textbf{(a)} KL divergence between the attention of HLA and of the full cache when the queries of a 2{,}048-token sequence are moved later by the given number of positions; thick lines average three RULER tasks, thin lines show each task. \textbf{(b)} RULER accuracy, with the full cache in orange.}
\label{fig:shift}
\end{figure}

Shifted positions were ablated on an earlier model with a window of 32 tokens, uptrained for 600 first-phase steps and then continued for 100 steps twice, with and without them (\cref{sec:app-shift}).
To change only the distance, we move the last 256 queries of a 2{,}048-token sequence up to 14K positions later, which lengthens every distance between a query and a key and leaves the hidden states unchanged.
Without shifted positions the divergence from the full cache rises from 0.022 to 0.098 at a shift of 4K, and with them only to 0.026 (\cref{fig:shift}a).
On RULER, where the distances come from real long prompts, shifted positions raise accuracy from 71.2 to 73.7 at 8K and from 54.3 to 62.7 at 16K (\cref{fig:shift}b), a gain that grows with the context length.

The long-document phase was evaluated on an earlier model with a window of 128 tokens: under the original RULER settings (\cref{sec:app-eval}), its accuracy at 16K rose from 69.6 before the phase to 71.2 after it.

%% file: sections/limitations.tex
\section{Limitations and Future Work}
\label{sec:limitations}

With the pretrained weights frozen, HLA does not fully match the full cache, and the gap grows with context length.
On RULER, Ouro-1.4B loses 0.8 points at 4K and 3.1 at 16K, where it keeps 96.0\% of its original accuracy (\cref{tab:quality}).
Contexts beyond 16K, where the smaller cache saves the most memory, are untested, and the gap may widen there.
HLA is fitted to a model that was pretrained with the full cache, so this gap measures how well new linear maps can imitate an existing model, and it need not be the limit of the design.
Pretraining a looped model with HLA from the start, so that its hidden states adapt to the latent, would separate the two.

HLA speeds up generation only, and a single sequence gains only beyond a few thousand tokens of context.
Prompts are processed with exact attention, so the prompt pass costs the same as with the full cache.
Each decoding step carries a fixed cost that the full cache does not have: a single sequence with 1K tokens of context decodes 26\% slower on Ouro-1.4B, and the two caches break even near 4K, although the larger batch already gives 2.5 times the throughput at 1K (\cref{sec:exp-efficiency}).
We attribute this cost to applying the maps and to attending to the window and the latent separately, which a kernel that merges the two could reduce.

Our evidence covers one family of looped models with $T{=}4$ loops at two sizes.
The main latent has the same size for any number of loops while the full cache grows with $T$, so the savings would grow with more loops, but whether a latent of this size can serve them is untested.
The window and the latent sizes were chosen on Ouro-1.4B (\cref{sec:app-sizes}) and reused for Ouro-2.6B without tuning.
Fine-tuning was tested on math reasoning only, and it runs two forward passes per step where standard fine-tuning runs one.
HLA changes what is stored per token, so it is complementary to methods that lower the precision of the cache or keep fewer tokens; we have not evaluated such combinations.

%% file: sections/conclusion.tex
\section{Conclusion}
\label{sec:conclusion}

A looped language model caches the keys and values of every loop, and this cache limits how many sequences a GPU decodes together.
Hybrid Latent Attention stores each older token as compact latents that every loop reads directly through its own linear maps, and keeps exact keys and values only for the most recent tokens.
Uptrained on Ouro-1.4B and Ouro-2.6B with frozen weights, it shrinks the cache by $10.7\times$ per token and raises peak decoding throughput by up to $7.4\times$ while retaining over 96\% of the original accuracy, and it remains trainable.
At the context lengths we test, a latent that each loop reads directly can thus replace most of the per-loop cache without reconstructing keys or values.

%% file: sections/appendix.tex
\section{Initialization of the Latent}
\label{sec:app-init}

This section details how the writer and the reader are initialized from principal components of the pretrained keys and values (\cref{sec:uptraining}), including the rotation $\tilde{R}_j$ of \cref{sec:reader} that keeps the key latent consistent with the rotary embeddings of the original model.

\paragraph{Frequency assignment.}
A head of dimension $d_h$ has $d_h/2$ rotary frequencies, one for each pair of key coordinates.
We write $k^f_{j,t,h} \in \mathbb{R}^2$ for the pair of frequency $f$ in the key of head $h$ at loop $t$; the rotation $R_j$ turns it into $\rho(j\theta_f)\, k^f_{j,t,h}$, where $\rho(\alpha)$ is the $2\times2$ rotation by $\alpha$.
The key latent has $r_k/2$ pairs, and the frequencies are assigned to them in round-robin order: pair $p$, counted from zero, receives frequency $p \bmod (d_h/2)$.
Each frequency thus receives $m = r_k/d_h$ latent pairs, and $\tilde{R}_j$ turns every one of them by $j\theta_f$.

\paragraph{Rotation commutes with same-frequency combinations.}
For one frequency $f$, we stack the key pairs of all heads and of loops 2 to $T$ as the rows of $K^f_j \in \mathbb{R}^{(T-1)H \times 2}$.
A combination matrix $M_f \in \mathbb{R}^{m \times (T-1)H}$ maps them to the $m$ latent pairs of that frequency, $C^f_j = M_f K^f_j$.
Rotation acts on every row from the right and $M_f$ mixes rows from the left, so
\begin{equation}
M_f \bigl(K^f_j\, \rho(j\theta_f)^\top\bigr) = \bigl(M_f K^f_j\bigr)\, \rho(j\theta_f)^\top.
\label{eq:rope-commute}
\end{equation}
Rotating the stored latent is therefore the same as writing it from rotated keys.
A latent pair that mixed key pairs of different frequencies would have no single angle with this property, which is why every latent pair carries one frequency.

\paragraph{Scores at initialization.}
The writer and the reader are initialized from one matrix $M_f$ with orthonormal rows per frequency.
The rows of $E_t$ for frequency $f$ are the combinations, with the coefficients of $M_f$, of the rows of the key projection, so that the key part of the latent is $C^f_j = M_f K^f_j$.
The map $A_t$ of head $h$ sends the query pair $q^f_{i,t,h}$ to the $m$ latent pairs of frequency $f$ with the coefficients $M_f[:,(t,h)]$, the column of $M_f$ that belongs to this head and loop.
Summing over the $m$ latent pairs, the contribution of frequency $f$ to the score $\hat{s}_{ij,t}$ of \cref{eq:latent-score} is, up to the factor $1/\sqrt{d_h}$,
\begin{equation}
\begin{gathered}
\sum_{(t',h')} \bigl[P_f\bigr]_{(t,h),(t',h')}\, \bigl(\rho(i\theta_f)\, q^f_{i,t,h}\bigr)^\top \bigl(\rho(j\theta_f)\, k^f_{j,t',h'}\bigr), \\
P_f = M_f^\top M_f,
\end{gathered}
\label{eq:rope-proj}
\end{equation}
where \cref{eq:rope-commute} moves the rotation from the latent to the keys.
$P_f$ is the orthogonal projection onto the $m$ directions spanned by the rows of $M_f$.
When $m = (T-1)H$, $P_f$ is the identity, and \cref{eq:rope-proj} summed over frequencies is the original score $s_{ij,t}$ of \cref{eq:exact}.
When $m < (T-1)H$, the key of head $h$ at loop $t$ is replaced by its row of $P_f K^f_j$, the projection of the stacked keys onto $m$ directions shared by all heads and loops of that frequency.

\paragraph{Choice of the directions.}
The rows of $M_f$ are the $m$ leading eigenvectors of the second-moment matrix $\Sigma_f = \sum_j K^f_j (K^f_j)^\top$, accumulated over the tokens of 128 calibration sequences from the reference model.
Among projections onto $m$ directions, this choice minimizes the squared error $\sum_j \lVert K^f_j - P_f K^f_j \rVert^2$ on these tokens.
Both coordinates of a pair enter $\Sigma_f$, so $\Sigma_f$ is unchanged when every pair is rotated by the same angle, and the same projection is optimal at every position.

\paragraph{Values and loop 1.}
Values carry no rotation.
The value part of the latent takes the $r_v$ leading principal components of the values of loops 2 to $T$ concatenated across heads and loops, and $B_t$ takes the block of these components that belongs to loop $t$.
The loop-1 latent is built in the same way from the $H$ key pairs of each frequency at loop 1, with $r_1/d_h$ latent pairs per frequency, and from the $r_1$ leading principal components of the loop-1 values.

\paragraph{Instantiation.}
Both models have heads of dimension $d_h{=}128$ and hence 64 frequencies, $H{=}16$ heads and $T{=}4$ loops.
With $r_k{=}512$, each frequency receives $m{=}4$ latent pairs out of $(T-1)H = 48$ key pairs; with $r_1{=}256$, the loop-1 latent keeps 2 of the $H{=}16$ key pairs of each frequency.

\section{Rotating the Whole Key Latent}
\label{sec:app-rope}

The construction of \cref{sec:app-init} rotates every pair of the key latent. This section describes what changes after uptraining, why HLA rotates the whole latent rather than a part of it, and what this choice costs.

\paragraph{After uptraining.}
Uptraining updates $E_t$, $A_t$ and $B_t$ as dense matrices, and only the assignment of frequencies to latent pairs stays fixed.
A latent pair of frequency $f$ can then also carry key components of other frequencies, which $\tilde{R}_j$ turns at the angle of $f$ rather than at their own, so \cref{eq:rope-proj} no longer holds exactly.
The scores still depend on the positions only through $i-j$, because the query and the latent are rotated in the same space, but their dependence on distance may differ from that of the original model.
Training with shifted positions keeps the learned scores accurate over long distances (\cref{sec:app-shift}).

\paragraph{Why the whole latent is rotated.}
MLA rotates only a small part of each key, a rotary key shared by the heads, and caches the rest without position~\citep{deepseek2024v2}; conversions of pretrained models to MLA adopt this split and drop the rotation from most key dimensions~\citep{meng2025transmla}.
HLA rotates every latent pair instead.
For frequency $f$, the original score depends on the positions only through $\rho(i\theta_f)^\top\rho(j\theta_f) = \rho((j-i)\theta_f)$.
A position-free component replaces this rotation by the identity, which changes its contribution to the score by $q^\top\bigl(\rho((j-i)\theta_f) - I\bigr)k$.
This term vanishes only when $(j-i)\theta_f$ stays close to zero over the distances in use or when the component carries no energy, and it remains at full rank.
The rotated latent has no such term: by \cref{eq:rope-proj}, its only approximation at initialization is the projection $P_f$.

In a looped model, the rotary part must also serve the keys of every loop.
A rotary key computed from one hidden state, as in MLA, approximates the keys of the other loops, which differ from it (\cref{tab:observations}), and one rotary key per loop makes this part of the cache grow with $T$.
A rotary key that combines the loops through \cref{eq:rope-commute} avoids both, and it is a rotated latent of smaller size.
The two designs thus differ only in whether part of the key budget drops the rotation, and HLA keeps the rotation on all of it.

\paragraph{Limitations of the rotated latent.}
This argument concerns the initialization; we have not compared the two designs after uptraining.
Rotating every pair also has two costs.
Every frequency receives the same number of latent pairs, whereas the eigenvalues of $\Sigma_f$ need not be spread evenly across frequencies, and a budget allocated per frequency by these eigenvalues could keep more of the original scores at the same rank.
After uptraining, every latent pair carries a rotation whose alignment with the original frequencies is no longer guaranteed, so the dependence of the scores on distance has to be learned.
Without shifted positions, the divergence from the original attention rises from 0.022 to 0.098 when the queries are moved 4K positions later (\cref{sec:app-shift}).
A position-free part of the latent would be insensitive to distance by construction.
A comparison at equal cache size between the rotated latent and a split into a rotary key and a position-free latent, and an allocation of pairs by frequency energy, are left to future work.

\section{Training Details}
\label{sec:appendix}

\paragraph{Uptraining.}
Both phases use AdamW with $\beta = (0.9, 0.95)$, weight decay 0.01 and gradient clipping at 1.0, with a batch of 128 sequences and a cosine schedule that decays the learning rate to one tenth of its peak.
The first phase uses a peak learning rate of $10^{-3}$ with 50 warmup steps, and the second $3\times10^{-4}$ with 10 warmup steps.
Sequences of the first phase have 64 to 2{,}048 tokens and mix math reasoning traces from OpenR1-Math-220k~\citep{openr1math2025} and web text from the 10B-token sample of FineWeb-Edu~\citep{penedo2024fineweb} at a ratio of 3:2.
The long documents of the second phase are windows of 16{,}384 tokens cut from the book and repository-level code subsets of the ProLong data~\citep{gao2025prolong}, 983 from books and 967 from code.
The loss on each of them is computed on 1{,}024 query positions.

\begin{figure}[!ht]
\centering
\includegraphics[width=\linewidth]{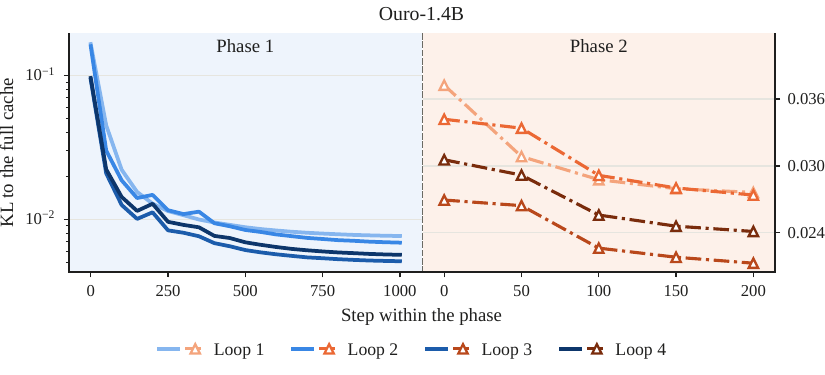}
\caption{Uptraining curves on held-out data for Ouro-1.4B, drawn as in \cref{fig:curves}.}
\label{fig:curves-14}
\end{figure}

\section{Where the Latent Errs}
\label{sec:app-distance}

This section details the measurement of \cref{fig:distance}, which motivates the exact window (\cref{sec:window}).

\paragraph{Setup.}
We use the HLA model without a window of \cref{tab:ablation}, uptrained on Ouro-1.4B with the full recipe.
The measurement takes 16 held-out math documents of 2{,}048 tokens and 32 evenly spaced query positions in each.
The original model is run on each document, and at every layer and loop we compute two attention distributions from its hidden states at 32-bit precision: the original one, $a^\star_{i\cdot,t}$, and that of HLA, $\hat{a}_{i\cdot,t}$.
Both start from the same hidden states, so their difference is the error of a single attention step, separate from errors that accumulate over layers.

\paragraph{Split by distance.}
The KL divergence of one query is a sum over the tokens it attends to,
\begin{equation}
\begin{aligned}
&\operatorname{KL}\!\left(a^\star_{i\cdot,t} \,\middle\|\, \hat{a}_{i\cdot,t}\right) \\
&= \sum_{j \le i} \left[ a^\star_{ij,t} \log \frac{a^\star_{ij,t}}{\hat{a}_{ij,t}} - a^\star_{ij,t} + \hat{a}_{ij,t} \right],
\end{aligned}
\label{eq:kl-split}
\end{equation}
and every term of this sum is nonnegative.
We assign the term of token $j$ to the distance $i-j$, and add up the terms and the original attention $a^\star_{ij,t}$ within six ranges of distance, over all queries, heads, layers and loops.

\paragraph{Result.}
\Cref{fig:distance} shows the two sums per range of distance, and their ratio.
Beyond the numbers given in \cref{sec:exp-ablation}, tokens at distances of 1{,}024 and more hold 29.5\% of the attention and 11.0\% of the divergence.
The pattern is the same in every loop, where distances 1 to 127 account for 47\% to 51\% of the divergence.

\section{Two-Pass Fine-Tuning}
\label{sec:app-twopass}

This section supports the two claims made about the procedure of \cref{sec:finetune}.

\paragraph{Repetition reaches sequential decoding.}
Consider a training sequence whose first $p$ tokens are the prompt.
In a parallel pass under \cref{eq:hybrid}, token $i$ reads the exact keys and values of the tokens in its window, the loop-1 latents of earlier tokens, and their main latents.
The first two are produced in the same pass, in the order in which decoding produces them, because a loop-1 latent depends only on the hidden state of its own loop.
Only the main latents are taken from the previous pass.
The hidden states of token $i$ therefore equal those of sequential decoding once the main latents of all tokens $j < i-W$ do.
Prompt tokens attend exactly in training and in decoding, so their latents are already those of decoding after the first pass.
The second pass is then exact for the first $W{+}1$ response tokens, which read main latents of prompt tokens only, and each repetition extends the exact range by another $W{+}1$ tokens.
A response of $m$ tokens is thus reproduced exactly after $\lceil m/(W{+}1) \rceil$ passes under the decoding rule.

\paragraph{Two passes are already close.}
The bound above counts the passes needed for an exact match.
To see how close fewer passes come, we take the uptrained HLA models of the main experiments and run 32 fine-tuning validation examples, with 47{,}933 response tokens, through each procedure and through sequential decoding under teacher forcing.
For each response token we compute the KL divergence from the next-token distribution of sequential decoding to that of the procedure, and average it over tokens (\cref{tab:twopass}).
At 32-bit precision, a single pass with exact attention differs from decoding by $5.6\times10^{-3}$ on Ouro-1.4B and $3.8\times10^{-3}$ on Ouro-2.6B.
The two-pass procedure lowers this by a factor of 16 to 19, and the first repetition by roughly another factor of ten.
At the 16-bit precision used for training, the two-pass procedure differs by $5.8\times10^{-4}$ and $4.0\times10^{-4}$, and one repetition only halves this, to about the level that 32-bit runs reach without it.
At this precision, rounding likely accounts for much of the remaining difference, so a repetition gains less.
Under the two-pass procedure the cross-entropy of the responses moves by at most $3\times10^{-4}$ nats at either scale and precision.

\begin{table}[!ht]
\centering
\caption{KL divergence ($\times10^{-4}$) from sequential decoding to each training procedure, averaged over the 47{,}933 response tokens of 32 fine-tuning validation examples, for the HLA models of the main experiments ($W{=}128$). Two repetitions were measured at 32-bit precision only.}
\label{tab:twopass}
\vspace{2pt}
\small
\begin{tabular}{lcccc}
\toprule
 & \multicolumn{2}{c}{Ouro-1.4B} & \multicolumn{2}{c}{Ouro-2.6B} \\
\cmidrule(lr){2-3}\cmidrule(lr){4-5}
Procedure & 32-bit & 16-bit & 32-bit & 16-bit \\
\midrule
Single exact pass & 56 & 58 & 38 & 40 \\
Two passes & 3.4 & 5.8 & 2.0 & 4.0 \\
One repetition & 0.29 & 2.5 & 0.19 & 2.0 \\
Two repetitions & 0.05 & -- & 0.02 & -- \\
\bottomrule
\end{tabular}
\end{table}

\paragraph{Settings.}
Repetitions are run without gradients, and the latents they produce keep the gradient of the first pass.
Ouro-1.4B is fine-tuned with one repetition.
For Ouro-2.6B, a repetition would lower the divergence at the 16-bit training precision only from $4.0\times10^{-4}$ to $2.0\times10^{-4}$ (\cref{tab:twopass}), and the deeper model is fine-tuned without it to limit training time.

\section{Shifted Positions}
\label{sec:app-shift}

This section details the ablation of shifted positions in \cref{sec:exp-ablation}.
It was run on an earlier configuration of Ouro-1.4B with a window of 32 tokens.

\paragraph{Setup.}
The starting point is an HLA model with the latent sizes of the main experiments and $W{=}32$, uptrained for 600 steps on the first-phase data without shifted positions.
We continue it twice under the same settings, with a peak learning rate of $3\times10^{-4}$: once with shifted positions as described in \cref{sec:uptraining}, and once without.
Both continuations are evaluated after 100 steps.

\paragraph{Attention under a position shift.}
The first measurement changes the distance between queries and keys while keeping their content fixed.
We take the hidden states of the original model on the first 2{,}048 tokens of 24 RULER prompts, eight from each of three tasks, and move the last 256 queries $D$ tokens later while the keys stay in place.
Every distance between a query and a key then grows by $D$, and no query, key or latent changes.
\Cref{fig:shift}(a) shows the KL divergence between the attention of the full cache and that of HLA over the tokens at least 128 positions before the query, averaged over layers and loops.
The two continuations agree at $D{=}0$, with a divergence of 0.022.
Without shifted positions the divergence rises to 0.098 at $D{=}4$K and stays near 0.09 up to 14K, whereas with shifted positions it is 0.026 at 4K and 0.032 at 14K.
The error of the model trained on short sequences thus comes from distances it has not seen, and shifted positions remove most of it without longer inputs.

\paragraph{RULER.}
\Cref{fig:shift}(b) compares the two continuations on RULER.
Shifted positions raise the average from 81.5 to 82.3 at 4K, from 71.2 to 73.7 at 8K and from 54.3 to 62.7 at 16K, so the gain grows with the context length, as the first measurement predicts.
On the eleven retrieval and aggregation tasks, the average rises from 58.7 to 72.8 at 16K.

\section{Latent and Window Sizes}
\label{sec:app-sizes}

The sizes $r_k{=}r_v{=}512$, $r_1{=}256$ and $W{=}128$ were chosen on Ouro-1.4B and reused for Ouro-2.6B.
This section reports the measurements behind this choice.

\paragraph{Latent size at initialization.}
The initialization of \cref{sec:app-init} can be evaluated before any training.
\Cref{tab:size-init} varies one rank at a time and reports the error on 64 held-out sequences.
The attention divergence of loops 2 to 4 depends on the key rank only, and the error of the attention output on the value rank only, so the two ranks can be set separately.
All three errors fall steadily with the rank and show no point beyond which a larger latent stops helping.
The errors are measured without a window, and uptraining reduces them by more than an order of magnitude (\cref{sec:exp-curves}), so this table shows what each rank controls and not the accuracy it leads to.

\begin{table}[!ht]
\centering
\caption{Error of the initialized latent on Ouro-1.4B when one rank varies and the others stay at $r_k{=}r_v{=}512$, $r_1{=}256$. KL: divergence between the attention of HLA and of the full cache, in nats, averaged over the loops the latent serves. Output error: relative squared error of the attention output, averaged over loops 2--4.}
\label{tab:size-init}
\vspace{2pt}
\small
\setlength{\tabcolsep}{4pt}
\begin{tabular}{rccc}
\toprule
 & Key rank $r_k$ & Value rank $r_v$ & Loop-1 rank $r_1$ \\
Rank & KL, loops 2--4 & Output error & KL, loop 1 \\
\midrule
128 & 1.04 & 0.42 & 1.04 \\
256 & 0.51 & 0.37 & 0.44 \\
512 & 0.29 & 0.30 & 0.23 \\
1{,}024 & 0.11 & 0.22 & 0.08 \\
2{,}048 & 0.03 & 0.16 & -- \\
\bottomrule
\end{tabular}
\end{table}

\paragraph{Latent size after uptraining.}
We then uptrained models of different sizes with the first phase, without shifted positions, and evaluated them on MATH500 with one sample per problem, where one standard error is about 2 points.
\Cref{tab:size-trained} gives the best checkpoint of each run.
Halving every rank to 256 loses about 15 points.
Doubling the key rank, or both ranks, to 1{,}024 changes accuracy by at most 3.4 points while the cache grows by a quarter.
A window of 32 tokens on the 72\,KiB latent reaches 71.8, above the model without a window whose latent is 1.7 times as large.
With this window, a loop-1 latent of rank 512 gives 70.6 against 71.8 at rank 256, a difference within one standard error.
We therefore kept $r_k{=}r_v{=}512$ and $r_1{=}256$ and spent the remaining memory on the window.

\begin{table}[!ht]
\centering
\caption{MATH500 accuracy (\%) of Ouro-1.4B after the first uptraining phase for different latent sizes, with one sample per problem.}
\label{tab:size-trained}
\vspace{2pt}
\small
\setlength{\tabcolsep}{3pt}
\begin{tabular}{cccrc}
\toprule
$r_k$ & $r_v$ & $r_1$ & Cache per token (KiB) & MATH500 \\
\midrule
\multicolumn{5}{l}{\emph{Without a window}} \\
256     & 256     & 256 &  48 & 51.4 \\
512     & 512     & 512 &  96 & 66.8 \\
1{,}024 & 512     & 256 &  96 & 66.6 \\
1{,}024 & 1{,}024 & 256 & 120 & 70.2 \\
\midrule
\multicolumn{5}{l}{\emph{With a window of 32 tokens}} \\
512     & 512     & 256 &  72 & 71.8 \\
512     & 512     & 512 &  96 & 70.6 \\
\bottomrule
\end{tabular}
\end{table}

\paragraph{Window size.}
A window of 32 tokens was our first choice, and the model of \cref{sec:app-shift} uses it.
Two observations led us to 128.
The divergence per unit of attention in \cref{fig:distance} is 0.049 nats at distances 32 to 127 against 0.040 at distances 1 to 31, so a window of 32 tokens leaves to the latent the range where it is least accurate.
On RULER at 16K, the model with $W{=}32$ also repeated the instruction that ends a question-answering prompt in 27 of 100 HotpotQA examples, against 1 for the full cache; the repeated segment starts a median of 54 to 61 tokens before the end of the prompt, outside a window of 32 tokens and inside one of 128.
A window of 128 tokens covers both and costs 96\,MiB per sequence on Ouro-1.4B.

\paragraph{Math and long-context retrieval.}
MATH500 separates these configurations much less than RULER does.
The model with $W{=}32$ scores 71.8, 71.6, 71.4, 70.0, 70.8 and 71.2 on MATH500 after 100 to 600 uptraining steps, so it reaches its level within the first 100 steps, about 4 points below the 75.9 of the full cache.
After 600 steps its RULER accuracy is 81.5, 70.3 and 54.2 at 4K, 8K and 16K, against 85.2, 80.7 and 74.7 for the full cache.
The same model thus keeps about 95\% of the original accuracy on math and 73\% on retrieval at 16K.
Shifted positions raise its accuracy at 16K to 62.7 (\cref{sec:app-shift}), and the main configuration reaches 69.1 under the same settings (\cref{sec:app-eval}).

\section{Evaluation Details}
\label{sec:app-eval}

\paragraph{Question answering in RULER.}
Two of the 13 RULER tasks, built from SQuAD~\citep{rajpurkar2016squad} and HotpotQA~\citep{yang2018hotpotqa}, ask a question about the given documents; lm-evaluation-harness ends their prompt with ``Answer:'' and generates at most 32 tokens.
The other 11 tasks end their prompt with the beginning of the answer sentence, so the model writes the answer first.
After the question-answering prompt, Ouro often starts its reply by reasoning about the question instead, and the 32 tokens end before it reaches an answer.
On Ouro-1.4B at 16K, 52\% of the SQuAD replies of the full cache start this way, and none of them contains the answer.
The score of a cache then depends on how often it leads the model into such a reply.
The LLA decoder never does so at 16K and scores 60.4 on SQuAD under these settings against 31.8 for the full cache, although on the prompts where the full cache answers directly the two reach 69.6 and 66.1.
For \cref{tab:quality} we therefore place each question-answering prompt in the chat template of Ouro, with its default system message, the prompt as the user turn and the assistant header after it, and generate up to 2{,}048 tokens until the end of the turn.
The answer is scored with the string match of RULER over the whole reply; at most 12\% of the replies of the full cache, LLA and HLA reach the token limit, against 36\% for single-loop sharing on Ouro-1.4B.
Each RULER entry of \cref{tab:quality} averages the other 11 tasks under their original settings with these two tasks.
With the original settings for all 13 tasks, the full cache scores 85.2, 80.7 and 74.7 at 4K, 8K and 16K on Ouro-1.4B and HLA 83.2, 78.7 and 69.1.
Every other RULER result in the paper uses the original settings (\cref{tab:ablation}, \cref{fig:shift}, the long-document comparison in \cref{sec:exp-ablation}, and \cref{sec:app-sizes,sec:app-serving}) and is compared only with results under the same settings.

\paragraph{LLA.}
The accuracy of LLA in \cref{tab:quality} uses a codec with 96 latent dimensions per head for keys and 160 for values, distilled as its authors describe.
Its projections start from the top singular vectors of the keys and values of all loops and are trained with the pretrained weights frozen, on the KL divergence of the output distribution when every key and value is replaced by its reconstruction plus a squared error on the attention outputs weighted by 0.5.
We train for 1{,}000 steps on the data of our first phase and keep the checkpoint with the lowest held-out divergence, 0.017 for Ouro-1.4B and 0.011 for Ouro-2.6B.
Decoding follows the two-pass decoder of LLA: each step first runs the original model on the new token and writes the latent of its keys and values, then recomputes the step with every key and value, those of the new token included, rebuilt from the latent, and samples from this second pass.
We implement this decoder in vLLM and compare it with a PyTorch implementation on 8 MATH500 prompts with 256 greedy tokens each, where the two choose the same token at 99.6\% of the positions.
The implementation also keeps the exact cache for the first pass, so it reproduces the accuracy of LLA but not its memory, and we use it only for \cref{tab:quality}.

\section{Serving Implementation}
\label{sec:app-serving}

HLA runs in vLLM~0.26 as a model class for Ouro that uses the engine's paged cache, scheduler and attention kernels; the engine is unchanged apart from logging of preemptions, which the measurements use.
This section describes how the cache and one decoding step are mapped onto it.

\paragraph{Cache layout.}
Each layer declares three kinds of cache to the engine.
The main latent is stored as a cache with a single key--value head of width $r_k$, whose key half holds $c^K_j$ and whose value half holds $c^V_j$, and the loop-1 latent as a second cache of width $r_1$.
All $H$ heads of the layer read these two caches, as the heads of a group read one key--value head in grouped-query attention.
The exact window uses one ordinary per-head cache for each loop, declared with a sliding window of $W{+}1$ tokens, so the engine frees the blocks that fall out of the window and the memory of this cache does not grow with the sequence.
The engine allocates the three kinds from one pool of blocks.

\paragraph{One decoding step.}
The $T$ loops of a token run inside one forward pass of the model.
In each layer and loop, the query is mapped by $A_t$ and rotated with the latent rotation, and a paged decoding kernel scores it against the latent keys of the tokens before the window and returns the weighted sum of their value latents together with the log-sum-exp of the scores.
A paged FlashAttention call reads the exact keys and values of the window and of the current token in the same way.
The two partial results are merged with their log-sum-exp, which is equal to the single softmax of \cref{eq:hybrid}, after $B_t$ has mapped the latent part back to the head dimension.
The writer accumulates its sum over the loops within the forward pass and stores the loop-1 latent after the first loop and the main latent after the last, with the key part rotated once at that point.
A decoding step contains no synchronization with the host and no loop over requests, so it is captured and replayed as a CUDA graph like the step of the original model.

\paragraph{Prompts.}
A prompt is processed in one pass with exact attention over its own tokens, and its latents and window are written from that pass.
Prefix caching and chunked prefill are disabled for both HLA and the full cache: with them, whether a prompt token is read exactly or through its latent would depend on how the scheduler splits the prompt.

\paragraph{Capacity.}
The capacity that the engine reports at startup budgets the sliding-window caches pessimistically and underestimates how many sequences fit.
The concurrent sequences of \cref{tab:efficiency} are therefore measured: we increase the batch until the scheduler no longer runs all its sequences together, and count a batch only if none of its sequences was preempted.

\paragraph{Agreement with the reference implementation.}
Uptraining and fine-tuning use a separate PyTorch implementation of \cref{eq:hybrid}.
We checked the serving path against it on RULER with the $W{=}32$ starting model of \cref{sec:app-shift}: the two implementations reach 81.5, 70.3 and 54.2 against 81.6, 70.4 and 54.2 at 4K, 8K and 16K.

\section{Decoding Measurements}
\label{sec:app-efficiency}

\Cref{tab:efficiency} gives the measurements plotted in \cref{fig:efficiency}.

\begin{table}[!ht]
\centering
\caption{Values of \cref{fig:efficiency}, measured with the same settings. LLA uses its reconstruct path with a latent of rank 256 per head, 96 for the keys and 160 for the values; since the codec does not affect the timing, it is a per-head PCA codec for Ouro-1.4B and a random one for Ouro-2.6B. Ratios compare HLA with the full cache; the best value of each group is in bold.}
\label{tab:efficiency}
\vspace{2pt}
\small
\setlength{\tabcolsep}{3pt}
\begin{tabular}{llrrrrrrrrrrr}
\toprule
 & & \multicolumn{4}{c}{Concurrent sequences} & \multicolumn{4}{c}{Throughput (tokens/s)} & \multicolumn{3}{c}{Latency (ms/token)} \\
\cmidrule(lr){3-6} \cmidrule(lr){7-10} \cmidrule(lr){11-13}
Model & Context & Full & LLA & HLA & Ratio & Full & LLA & HLA & Ratio & Full & LLA & HLA \\
\midrule
\multirow{4}{*}{Ouro-1.4B}
 & 1K  & 68 & 264 & \textbf{298} & $4.4\times$ & 1{,}305 & 93.6 & \textbf{3{,}231} & $2.5\times$ & \textbf{13.6} &  28.5 & 17.1 \\
 & 4K  & 20 &  77 & \textbf{139} & $7.0\times$ &   373 & 27.8 & \textbf{1{,}429} & $3.8\times$ & \textbf{20.6} &  54.1 & 20.8 \\
 & 8K  & 10 &  40 &  \textbf{84} & $8.4\times$ &   192 & 14.3 &   \textbf{797} & $4.2\times$ & 30.1 &  90.1 & \textbf{24.2} \\
 & 16K &  5 &  19 &  \textbf{40} & $8.0\times$ &    97 &  7.2 &   \textbf{428} & $4.4\times$ & 49.3 & 158.7 & \textbf{25.1} \\
\midrule
\multirow{4}{*}{Ouro-2.6B}
 & 1K  & 32 & 123 & \textbf{127} & $4.0\times$ &   519 & 46.3 & \textbf{1{,}278} & $2.5\times$ & \textbf{27.0} &  56.8 & 34.3 \\
 & 4K  &  9 &  36 &  \textbf{68} & $7.6\times$ &   147 & 13.7 &   \textbf{593} & $4.0\times$ & \textbf{41.0} & 107.5 & 41.5 \\
 & 8K  &  4 &  18 &  \textbf{35} & $8.8\times$ &    66 &  7.0 &   \textbf{327} & $5.0\times$ & 59.9 & 181.8 & \textbf{48.1} \\
 & 16K &  2 &   8 &  \textbf{13} & $6.5\times$ &    21 &  3.6 &   \textbf{151} & $7.4\times$ & 98.0 & 322.6 & \textbf{50.0} \\
\bottomrule
\end{tabular}
\end{table}